# DOSE PREDICTION DRIVEN RADIOTHERAPY PARAMETERS REGRESSION VIA INTRA- AND INTER-RELATION MODELING

*Jiaqi Cui[1], Yuanyuan Xu[1], Jianghong Xiao[2], Yuchen Fei[1], Jiliu Zhou[1], Xingcheng Peng[3,\*], Yan Wang[1,\*]*

[1]School of Computer Science, Sichuan University, China
[2]Department of Radiation Oncology, Cancer Center, West China Hospital, Sichuan University, China
[3]Department of Biotherapy, Cancer Center, West China Hospital, Sichuan University, China

**ABSTRACT**

Deep learning has facilitated the automation of radiotherapy by predicting accurate dose distribution maps. However, existing methods fail to derive the desirable radiotherapy parameters that can be directly input into the treatment planning system (TPS), impeding the full automation of radiotherapy. To enable more thorough automatic radiotherapy, in this paper, we propose a novel two-stage framework to directly regress the radiotherapy parameters, including a dose map prediction stage and a radiotherapy parameters regression stage. In stage one, we combine transformer and convolutional neural network (CNN) to predict realistic dose maps with rich global and local information, providing accurate dosimetric knowledge for the subsequent parameters regression. In stage two, two elaborate modules, i.e., an intra-relation modeling (Intra-RM) module and an inter-relation modeling (Inter-RM) module, are designed to exploit the organ-specific and organ-shared features for precise parameters regression. Experimental results on a rectal cancer dataset demonstrate the effectiveness of our method.

## 1. INTRODUCTION

Radiation therapy (RT) is one of the mainstay treatments for cancer in clinic [1]. In clinic, to reduce the radiation side effects of RT, a clinically acceptable RT plan is often necessitated before performing RT. To obtain a high-quality RT plan, dosimetrists have to manually set a series of radiotherapy parameters entered into the treatment planning system (TPS) and optimize them iteratively, due to the complex requirement of delivering therapeutic high doses to the planning target volume (PTV) while minimizing the doses to the organs at risk (OARs). However, since the parameters are set empirically, the whole process is operator-dependent and time-consuming, which may lead to delayed and compromised therapy. Therefore, it is of great interest to reduce manual interventions and accelerate RT planning.

In recent years, deep learning [2-4] has made a quantum leap in accelerating radiotherapy planning [5-14]. For example, Liu *et al.* [5] developed a U-ResNet-D framework for 3D dose prediction. More recently, Zhan *et al.* [7] constructed a Mc-GAN that employs an embedded UNet-like structure as a generator and gained impressive prediction accuracy. Despite their remarkable performance, the above methods ubiquitously cease in predicting only the dose distribution map, and the statistical dosimetric metrics (e.g., $V_{50}$, $D_{mean}$, and $D_{max}$) calculated from the predicted dose maps cannot be directly applied in the TPS. In other words, the dosimetrists are still required to manually translate the predicted dose maps to the appropriate radiotherapy parameters, which costs considerable time and suffers subjective expertise. Therefore, we claim that these methods which take the dose map prediction as the ultimate goal have not fully realized the automation of RT planning. Meanwhile, predicting the radiotherapy parameters that can be directly entered into the TPS holds clinical significance. Taking the above concerns into consideration, we propose to regress the desired radiotherapy parameters for further automating the RT planning. *To our knowledge, our work marks the pioneering effort in regressing accurate radiotherapy parameters for generating high-quality RT plans.*

Motivated to address the above limitations and facilitate a more thorough automated radiotherapy, in this paper, we innovatively propose a two-stage framework, including a dose map prediction stage to predict realistic dose maps from both computed tomography (CT) images and masks of PTV and OARs, and a radiotherapy parameters regression stage to accurately regress the radiotherapy parameters. Specifically, in stage one, given that the doses received by different organs are mutually affected, we introduce transformer into convolutional neural network (CNN) framework to capture such global dosimetric context while preserving local details in the predicted dose map. The predicted realistic dose map can provide accurate dosimetric information for the subsequent radiotherapy parameters regression. In stage two, we utilize the dosimetric information from the predicted dose map as well as the anatomical information from the CT

---
The first two authors contributed equally.
\*Corresponding author: Xingcheng Peng and Yan Wang.
Email: pxx2014@scu.edu.cn, wangyanscu@hotmail.com.

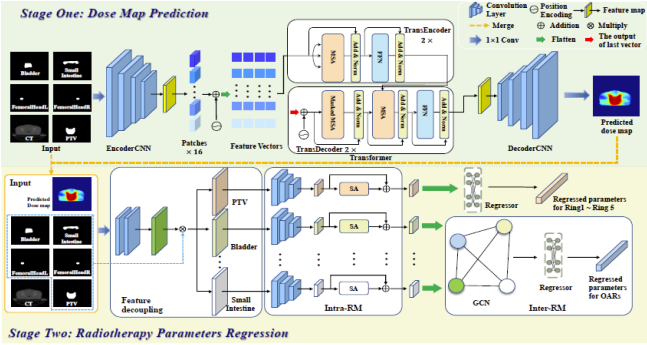

**Fig. 1.** Schematic view of our proposed method.

images to regress the radiotherapy parameters that can be directly applied for high-quality RT plan generation. Since PTV and OARs should be delivered to opposite doses and thus corresponding to different parameters, we first decouple them in the feature space. Then, we develop a self-attention-based intra-relation modeling (Intra-RM) module to help the decoupled organ-specific features capture the varying dose distributed on a certain organ within individual PTV and OARs. Meanwhile, we also develop a graph convolutional network (GCN)-based inter-relation modeling (Inter-RM) module to enable the interaction among OARs features and make them perceive the dose distributions among OARs. Through the two modules, the organ-specific features can be enhanced to better express the dose characteristics within and between different organs. Finally, these more expressive features are used to predict precise radiotherapy parameters of PTV and OARs via regressors. Experimental results demonstrate the superiority of our method.

## 2. METHODOLOGY

The overview of the proposed method is illustrated in Fig. 1, which consists of a dose map prediction stage and a radiotherapy parameters regression stage. Details of these two stages are elaborated in the following sub-sections.

### 2.1 Stage One: Dose Map Prediction

To produce a realistic dose map, two factors are supposed to be taken into account: 1) the local detailed information, such as the radiation edge within different organs and the surrounding x-ray penetrated area; 2) the global dosimetric context such as the mutually affected dose information among different PTV and OARs due to the complex and precise RT principles. Therefore, we embed the transformer into a UNet-like CNN framework, forming a network consisting of 1) a CNN-based encoder (EncoderCNN) to extract sufficient local details, 2) a transformer to capture the global dosimetric information, and 3) a CNN-based decoder (DecoderCNN) to restore the features obtained by the transformer and generate accurate dose map. Specifically, taking the original CT images and the corresponding masks of PTV and OARs as input, the EncoderCNN which contains 4 convolution blocks structured as $3 \times 3$ Convolution-BatchNorm-ReLU extracts the feature $f_{enc}$ with the local spatial details faithfully preserved. Given that transformer demands sequential input, we divide the $f_{enc}$ into $N$ patches and flatten them to obtain a 1D feature sequence $f_{seq}$. Then, to retain the spatial location information among patches, a position encoding is added to $f_{seq}$. The transformer that follows a standard architecture [15] receives the combined result $f_{seq}$ to capture the long-range dependencies among the mutually interacted organs, thus injecting rich global dosimetric context into the predicted dose map. The output sequence of transformer is reshaped to the feature $f_{dec}$ that shares the same size as $f_{enc}$. Finally, $f_{dec}$ is fed to DecoderCNN for generating the output of stage one, i.e., the predicted dose map $\hat{y}$. To speed up the convergence, we adopt the Huber loss to constrain the prediction $\hat{y}$ to be close to the corresponding ground truth (i.e., the clinically approved dose map) $y$, which can be formulated as follows:

$$L_{predict}(y,\hat{y}) = \begin{cases} \frac{1}{2(y-\hat{y})^2}, & |y-\hat{y}| < \delta \\ \delta\left(\left|y-\hat{y}-\frac{1}{2}\delta\right|\right), & |y-\hat{y}| \geq \delta \end{cases} \quad (1)$$

where $\sigma$ is the hyper-parameter and is empirically set to 0.5.

### 2.2 Stage Two: Radiotherapy Parameters Regression

Clinical RT planning is a complex and delicate task. Table 1 displays the radiotherapy parameters for a rectal cancer patient in clinic. As can be seen, to ensure that the dose received by PTV is sufficient but does not exceed the prescribed dose, the dosimetrists specify five dose-limiting rings around the PTV (i.e., Ring1-Ring5) and determine the maximum received dose for each ring, i.e., maxDose which is calculated from the parameter "weight" and "dose". Meanwhile, the dosimetrists need to decide the maximum dose delivered to a specific volume within each OAR, i.e., maxDVH which is derived by the parameter "weight", "volume", and "dose", to minimize the dose deposition in OARs. Besides the parameter "dose", other essential parameters, i.e., "weight" and "volume", cannot be directly regressed from dose distribution maps, namely, dosimetrists are still required to set and optimize them manually and iteratively, which is time-consuming and laborious. In our work, we would like to realize the accurate prediction of these radiotherapy parameters based on the high-quality dose map predicted in stage one. To achieve this, we construct another network in stage two, which takes the clinical images (i.e., CT and masks of PTV and OARs) and the predicted dose map in stage one as input, and outputs the desirable parameters.

***Feature Decoupling.*** The network first utilizes two convolution layers to extract the integrated feature $f_{int}$ that contains both the anatomical information from CT images and dosimetric knowledge from the predicted dose maps. Then, considering that each PTV and OAR corresponds to different parameters, we decouple $f_{int}$ into a set of organ-specific features, including the PTV features $f_{PTV}$ and OARs features $\{f_{OAR}^i\}_{i=1}^M$, where $M$ denotes the number of OARs. The feature decoupling is achieved by element-wise multiplying $f_{int}$ with the corresponding PTV and OARs masks, after which the organ-specific features are obtained.

**Table 1.** Example of parameters for a rectal cancer patient.

|  |  | Function Type | weight (%) | volume (%) | dose (Gy) |
|---|---|---|---|---|---|
| PTV | Ring1 | MaxDose | 20 | - | 48.38 |
|  | Ring2 | MaxDose | 20 | - | 45.00 |
|  | Ring3 | MaxDose | 20 | - | 86.00 |
|  | Ring4 | MaxDose | 20 | - | 40.32 |
|  | Ring5 | MaxDose | 20 | - | 28.22 |
| OARs | Bladder | MaxDVH | 20 | 31 | 40.00 |
|  | ST | MaxDVH | 20 | 22 | 30.00 |
|  | FHL | MaxDVH | 10 | 5 | 38.00 |
|  | FHR | MaxDVH | 10 | 5 | 38.00 |

**Intra-RM Module.** For individual PTV and OARs, the dose distribution varies with different intra regions Therefore, an Intra-RM module is proposed to obtain precise parameters within PTV and OARs. Particularly, the self-attention scheme is employed to calculate weights between every two anatomical regions (i.e., organs) in dose maps, thus modeling the dose distribution within organ-specific features. After the Intra-RM module, the enhanced PTV feature contains precise dosimetric information and can be used to predict the parameters for the five dose-limiting rings of PTV, denoted as $\{Pre_{PTV}^i\}_{i=1}^M$, through a fully connected (FC) layer-based regressor. Herein, we employ the mean absolute error (MAE) to calculate the loss between the predicted parameters of five dose-limiting rings $\{Pre_{PTV}^i\}_{i=1}^M$ and their corresponding ground truths $\{GT_{PTV}^i\}_{i=1}^M$ as follows:

$$L_{reg\_PTV} = \sum_{i=1}^{M} ||Pre_{PTV}^i - GT_{PTV}^i||_1 \quad (2)$$

**Inter-RM Module.** As for the OARs features, considering that clinical radiotherapy planning depends not only on the dose distribution within each OAR, but also on the inter-relation among OARs, we further develop an Inter-RM module to mimic such relationship. The Inter-RM module leverages a two-layer GCN [16] to model the correlations between OAR features, thus allowing them to interact with each other and perceive the dose distributions by learning the edges of the GCN. Concretely, before passing the Intra-RM-enhanced OARs features $\{f_{OAR}^i\}_{i=1}^M$ to the Inter-RM module, we first flatten $\{f_{OAR}^i\}_{i=1}^M$ into 1D vectors $\{\tilde{f}_{OARs}^i\}_{i=1}^M$. Each $\tilde{f}_{OARs}^i$ is reviewed as a node feature which is updated by integrating the features from other nodes. After each node is updated, the extracted more expressive features are input to a simple FC regressor to obtain the radiotherapy parameters of each OAR, denoted as $\{Pre_{OARs}^i\}_{i=1}^M$. Similarly, MAE loss is also adopted to encourage the similarity between the predicted radiotherapy parameters and the corresponding ground truths of OARs as follows:

$$L_{reg\_OARs} = \sum_{i=1}^{M} ||Pre_{OAR}^i - GT_{OAR}^i||_1 \quad (4)$$

where $GT_{OAR}^i$ represents the ground truth vector for $i$-th OAR.

In sum, the total loss function of the stage two is defined as follows:

$$L_{total} = L_{reg\_PTV} + \lambda L_{reg\_OARs} \quad (5)$$

where $\lambda$ is the only hyper-parameter and is set to 1.

### 2.3 Training Details
The proposed model is implemented on an NVIDIA GTX S2080s GPU with 8 GB memory. Both two stages are trained for 200 epochs with Adam optimize using a batch size of 4. The stage-one network is frozen during the stage-two training. The learning rate is set to 1e-5 for stage one and 1e-4 for stage two, respectively. The number $M$ of OARs is 4 for rectal cancer, which includes Bladder, SmallIntestine (ST), FemoralHeadR (FHR) and FemoralHeaL (FHL). Detailed training process of our model is presented in Algorithm 1.

---

**Algorithm 1:** The algorithm of the proposed model. *Net4Dose* denotes the dose prediction network in stage one with parameter $\alpha$. *Net4Params* is the radiotherapy parameters regression network with parameter $\beta$.

**Stage 1: Dose prediction**
**Input:** $x$ is the input including CT image and masks of PTV and OARs.
**Output:** $\hat{y}$ is the predicted dose distribution map, $y$ is the corresponding ground truth.
**while** epoch $i$ in total_epochs **do**
  $\hat{y}$=Net4Dose($x$)
  loss$^{dose}$ = huber_loss($\hat{y}$, $y$)
  $\alpha_i \xleftarrow{Adam} \alpha_{i-1}$
**end**

**Stage 2: RT Parameters Regression**
**Input:** $x$ and $\hat{y}$
**Output:** $\hat{z}$ and $z$ are the predicted RT parameters, $z$ is the corresponding ground truth.
**while** epoch $j$ in total_epoches **do**
  $\hat{z}$=Net4Params($x$, $\hat{y}$)
  loss$^{param}$ = MAELoss($z$, $\hat{z}$)
  $\beta_i \xleftarrow{Adam} \beta_{i-1}$
**end**

---

## 3. EXPERIMENTS AND RESULTS
### 3.1 Dataset and Evaluation Metrics
***Dataset.*** We evaluate the proposed method on an in-house rectal dataset including 131 patients. Each subject includes a CT image, a dose distribution map, masks of PTV and four OARs, and a set of radiotherapy parameters. The dose of the PTV is prescribed to 50.40Gy/28 fractions All the radiotherapy plans are conducted on the Raystation v4.7 TPS with the model for the Elekta Versa HD linear accelerators and are clinically approved. Meanwhile, the TPS parameters collected from the corresponding patients are also designed and approved by dosimetrists in clinic. Here, we randomly select 108 patients as the training set and the remaining 23 patients as the testing set. Each 3D CT image is divided into multiple 2D slices with a resolution of 512×512.

***Evaluation Metrics:*** To assess the quality of the predicted dose map in stage one, we adopt three commonly used criteria, including $V_{40}$, $V_{50}$, $D_{mean}$, $D_{max}$, conformity index (CI) [17], homogeneity index (HI) [18], and dose volume histogram (DVH). Furthermore, we calculate the differences |Δ| between predicted radiotherapy parameters and their ground truths (GT) to assess the accuracy of the estimation in stage two. The above metrics are calculated at the patient level.

### 3.2 Ablation Experiment
To explore the contributions of key components in the proposed model, we construct the following ablation variants: (A) w/o transformer; (B) w/o Intra-RM; (C) w/o Inter-RM; (D) w/o Intra-RM and Inter-RM; (E) proposed. The quantitative results are shown in Table 2. *(1) Effectiveness of transformer*: By comparing (A) and (E), we can find that the introduction of the transformer increases the MAD for CI and HI in stage one by 0.013 and 0.046, respectively. *(2) Effectiveness of Intra-RM and Inter-RM*: By comparing the results of (B) and (E), we can find that the removal of Intra-RM rises the |Δ| by 0.836% (weight) on Ring1 and 1.081Gy (dose) on Bladder. As for Inter-RM, we find that it can boost the performance by 2.123% (weight) on Ring 1 and 3.32 Gy (dose) and 1.279% (volume) on Bladder by comparing (C) and (E). Moreover, the performance exhibits a larger degradation when we remove both the Inter-RM and Intra-RM in (D). All results indicate the effectiveness of Intra- and Inter-RM modules in helping regress accurate parameters.

**Table 2.** Quantitative ablation results in terms of PTV CI and HI for stage one (left) and the predicted error |Δ| for stage two (right).

|     | Stage one |       | Stage two      |          |                |          |           |
| --- | --------- | ----- | -------------- | -------- | -------------- | -------- | --------- |
|     | CI        | HI    | Ring 1 maxDose |          | Bladder maxDVH |          |           |
|     | MAD       | MAD   | weight(%)      | dose(Gy) | weight(%)      | dose(Gy) | volume(%) |
| (A) | 0.059     | 0.061 | 1.665          | 3.371    | 0.840          | 2.482    | 3.493     |
| (B) |           |       | 1.027          | 2.387    | 0.441          | 1.785    | 1.178     |
| (C) |           |       | 2.314          | 2.994    | 0.765          | 2.884    | 1.922     |
| (D) |           |       | 3.002          | 3.002    | 1.553          | 1.633    | 5.365     |
| (E) | 0.046     | 0.015 | **0.191**      | **0.582**| **0.351**      | **0.691**| **0.643** |

### 3.3 Comparison with state-of-the-art methods

**Stage One.** we compare our model with four state-of-the-art (SOTA) dose prediction models, i.e., U-net [16], DoseNet [20], U-ResNet-D [5], and DeeplabV3+ [21]. The quantitative results in Table 3 show that our method (stage one) outperforms other methods in almost all criteria. Also, the $D_{mean}$ acquired by our method approximates the optimal one, with a tolerable difference of 0.161. Meanwhile, our method has far less parameters (19.21M) compared to the second-best performer DeeplabV3+ (56.62M). In addition, we provide visualizations in Fig. 2 where the dose distribution map predicted by our model yields the best visual effect (a) and the best matching DVH curves (b). Qualitatively and quantitatively, our method is proven to generate dose distribution maps of superior quality.

**Stage Two.** Given that this is the first work for radiotherapy parameters regression, we consider adapting the regression models in other fields, i.e., Regression CNN [22], MOCNN [23], to our task for comparison. Table 4 shows the quantitative results on parameters "weight", "volume" and "dose" for PTV as well as "weight" and "dose" for OARs. We can observe that the parameters regressed from our model obtain the minimum |Δ| from the real in most cases. Particularly, compared with Regression CNN, our method

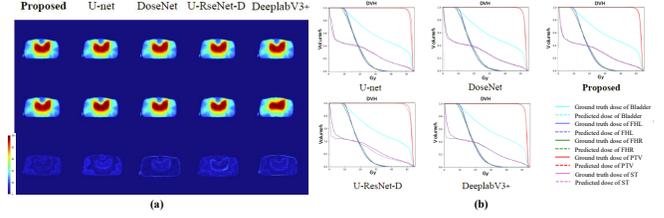

**Fig. 2.** Visualization results (a) and DVH plots (b) for comparison with dose prediction methods.

**Table 3.** Quantitative comparison with dose prediction methods in terms of $V_{40}$, $V_{50}$, $D_{mean}$, $D_{max}$. GT denotes the ground truth.

| Methods    | $V_{40}$      | $V_{50}$      | $D_{mean}$      | $D_{max}$       | Params |
| ---------- | ------------- | ------------- | --------------- | --------------- | ------ |
| UNet       | 0.964(0.157)  | 0.794(0.229)  | 49.032(5.279)   | 51.602(0.307)   | **0.59** |
| DoseNet    | 0.965(0.155)  | 0.793(0.174)  | **48.694(5.453)** | 51.684(0.230) | 5.73   |
| U-ResNet-D | 0.956(0.159)  | 0.689(0.213)  | 48.398(5.580)   | 51.805(0.273)   | 14.04  |
| DeeplabV3+ | 0.965(0.153)  | 0.798(0.180)  | 48.884(5.367)   | 51.778(0.216)   | 56.62  |
| Proposed   | **0.965(0.154)** | **0.467(0.176)** | 48.697(5.058) | **51.850(0.219)** | 19.21 |
| GT         | 0.962(0.165)  | 0.492(0.264)  | 48.536(5.849)   | 52.366(0)       | -      |

enhances the performance by 1.722% (weight) and 2.648 (dose) on Ring 1. Meanwhile, an enhancement of 1.632% (weight) on Ring2 and 1.369 (dose) on Ring 4 is achieved by our method, compared with MOCNN. For OARs, the proposed model greatly lowers the |Δ| of FHL by 5.701% and 0.813% (volume) compared to Regression CNN and MOCNN, respectively. Besides, p-values of the paired t-test are mostly less than 0.05, showing that our improvements are statistically significant. All these results suggest the advanced performance of our method in regressing precise radiotherapy parameters to facilitate more thorough radiotherapy.

## 4. CONCLUSION

In this work, to achieve more thorough automation of radiotherapy planning, we propose a two-stage model consisting of a dose map prediction stage and a radiotherapy parameters regression stage. To produce high-quality dose maps, we introduce the transformer into a CNN framework in stage one to fully capture both the local and global information. In stage two, the collaborative Intra-RM and Inter-RM are proposed to explore the varying dose distribution weight within independent PTV and OARs and capture the clinical correlations among OARs, respectively. Experimental results demonstrate the feasibility and superiority of our method on the rectal cancer dataset. Future work involves extending our method to the radiotherapy parameters prediction of other types of cancers.

**Table 4.** Quantitative comparison with regression methods in terms of maxDose (i.e., weight and dose) for PTV and maxDVH (i.e., weight, volume and dose) for OARs.

|      |         | Regression CNN |Δ| |           |          | MOCNN |Δ| |           |          | **Proposed (Stage two) |Δ|** |           |          |
| ---- | ------- | --------- | --------- | -------- | --------- | --------- | -------- | ----------- | ----------- | ---------- |
|      |         | weight(%) | volume(%) | dose(Gy) | weight(%) | volume(%) | dose(Gy) | weight(%)   | volume(%)   | dose(Gy)   |
| PTV  | Ring 1  | 3.043     | -         | 5.130    | 1.193     | -         | 3.371    | **1.321**   | -           | **2.482**  |
|      | Ring 2  | 1.451     | -         | 4.003    | 3.065     | -         | 2.387    | **1.442**   | -           | **1.785**  |
|      | Ring 3  | 1.130     | -         | 4.370    | 2.321     | -         | 2.994    | **0.861**   | -           | **2.884**  |
|      | Ring 4  | 1.292     | -         | 3.681    | 2.180     | -         | 3.002    | **0.884**   | -           | **1.633**  |
|      | Ring 5  | 1.524     | -         | 2.265    | **0.122** | -         | 2.393    | 0.412       | -           | **1.031**  |
| OARs | Bladder | 1.395     | 2.325     | 10.04    | 0.806     | 0.913     | 5.602    | **0.031**   | **0.803**   | **3.880**  |
|      | ST      | 1.345     | 1.593     | 8.710    | 0.543     | 0.229     | 4.595    | **0.063**   | **0.162**   | **1.891**  |
|      | FHR     | 4.013     | 3.350     | **1.177**| **0.831** | 0.845     | 4.281    | 3.340       | **0.174**   | 4.376      |
|      | FHL     | 3.281     | 6.472     | 8.181    | 2.030     | 1.584     | 3.983    | **0.812**   | **0.771**   | **2.832**  |

**COMPLIANCE WITH ETHICAL STANDARDS**

All procedures performed in studies involving human participants were in accordance with the ethical standards of the institutional and/or national research committee and with the 1964 Helsinki declaration and its later amendments or comparable ethical standards. This article does not contain any studies with animals performed by any of the authors.

**ACKNOWLEDGMENTS**

This work is supported by the National Natural Science Foundation of China (NSFC 6237132, 62071314), Sichuan Science and Technology Program 2023YFG0263, 2023YFG0025, 2023NSFSC0497, and Opening Foundation of Agile and Intelligent Computing Key Laboratory of Sichuan Province.